\documentclass{article}
\usepackage{ijcai24}

\usepackage{times}
\usepackage{soul}
\usepackage{url}
\usepackage[hidelinks]{hyperref}
\usepackage[utf8]{inputenc}
\usepackage[small]{caption}
\usepackage{graphicx}
\usepackage{amsmath}
\usepackage{amsthm}
\usepackage{booktabs}
\usepackage{algorithm}
\usepackage{algorithmic}
\usepackage[switch]{lineno}
\usepackage{mathtools}

\usepackage{xcolor}
\usepackage{amsfonts}
\usepackage{subcaption}
\usepackage{multirow}

\title{On the Feasibility of Fidelity$^{-}$ for Graph Pruning}

\author{
Yong-Min Shin$^1$\and
Won-Yong Shin$^{1,}$\footnote{Corresponding author.}
\affiliations
$^1$Yonsei University, Seoul, South Korea\\
\emails
\{jordan3414, wy.shin\}@yonsei.ac.kr
}

\begin{document}

\maketitle

\begin{abstract}
As one of popular quantitative metrics to assess the quality of explanation of graph neural networks (GNNs), \textit{fidelity} measures the output difference after removing unimportant parts of the input graph. Fidelity has been widely used due to its straightforward interpretation that the underlying model should produce similar predictions when features deemed unimportant from the explanation are removed. This raises a natural question: ``Does fidelity induce a global (soft) mask for graph pruning?" To solve this, we aim to explore the potential of the fidelity measure to be used for graph pruning, eventually enhancing the GNN models for better efficiency. To this end, we propose \textbf{F}idelity$^-$-\textbf{i}nspired \textbf{P}runing (\textbf{FiP}), an effective framework to construct global edge masks from local explanations. Our empirical observations using 7 edge attribution methods demonstrate that, surprisingly, general eXplainable AI methods outperform methods tailored to GNNs in terms of graph pruning performance. 
\end{abstract}

\section{Introduction}


Alongside the recent popularity of graph neural networks (GNNs) for graph-related tasks spanning across domains from social network recommendations~\cite{Wu2023recommendationsurvey} to molecular property predictions~\cite{Reiser2022molecularsurvey}, such developments have resulted in an increasing demand in developing eXplainable AI (XAI) methods for GNN models. While early work focused on extending various edge attribution methods into GNNs~\cite{Baldassarre2019earlygnnxai,Pope2019earlygnnxai} for explaining the underlying model's behavior, many XAI methods specifically designed with GNN models in mind have been since proposed~\cite{Yuan2023GNNXAISurvey}, e.g., GNNExplainer~\cite{Ying2019GNNExplainer}. More recent studies include FastDnX~\cite{Pereira2023FastDnX}, relying on training SGC~\cite{Wu2019SGC} as a simpler surrogate model to the original GNN to extract relevant subgraphs. Although there are alternative forms of explanations for GNN models, the most prevalent one lies in the form of locally identifying the most relevant subgraph structure to the GNN's output for a given node (i.e., the predicted node class).

One of the broader objectives of XAI is to ultimately enhance the performance based on the knowledge gained from the explanation~\cite{Samek2019,Ali2023XAIsurvey}. In this regard, even though the majority of XAI methods of GNNs have successfully developed effective explanations, studies on utilizing such explanations to \textit{improve} the underlying GNN model have been vastly underexplored. In the context of to graph datasets and GNN models, we focus on the problem of \textit{graph pruning}, which is related to increasing the GNN model's efficiency by removing unimportant edges from the underlying graph. In other words, we are interested in removing edges from the input graph altogether, guided by edge attributions from XAI methods. If such utilization of XAI is shown to be successful, then we are capable of naturally boosting the efficiency of the underlying GNN model, since the time complexity of most GNN models is directly determined by the number of input edges~\cite{Wu2021gnnsurvey}.

Our work aims at making a connection between graph pruning and \textit{fidelity}, a quantitative metric that is often used to assess the quality of (graph) explanations~\cite{Ancona2017fidelity,Yeh2019fidelity,Yuan2023GNNXAISurvey}. In the context of GNN explanations, two variants of fidelity are commonly used. First, fidelity$^{-}$ measures the output difference between two instances when the original input graph is used and when the `unimportant' parts (i.e., edges) are removed from the input graph. The intuition behind this metric is quite straightforward: if the explanation is valid, then structures deemed less important (i.e., assigned low edge attribution scores) should have less impact to the model's output after removal from the input. Second, for fidelity$^{+}$, the definition and interpretations are vice versa (i.e., removing `important' parts). Revisiting on the intuition of the fidelity$^{-}$ measure, we hypothesize that frequently removed edges, when measuring fidelity$^{-}$, may potentially be simply removed from the original graph, provided that the quality of the given explanation is good enough. 


\begin{figure*}[ht]
    \centering
    \includegraphics[width=0.9\linewidth]{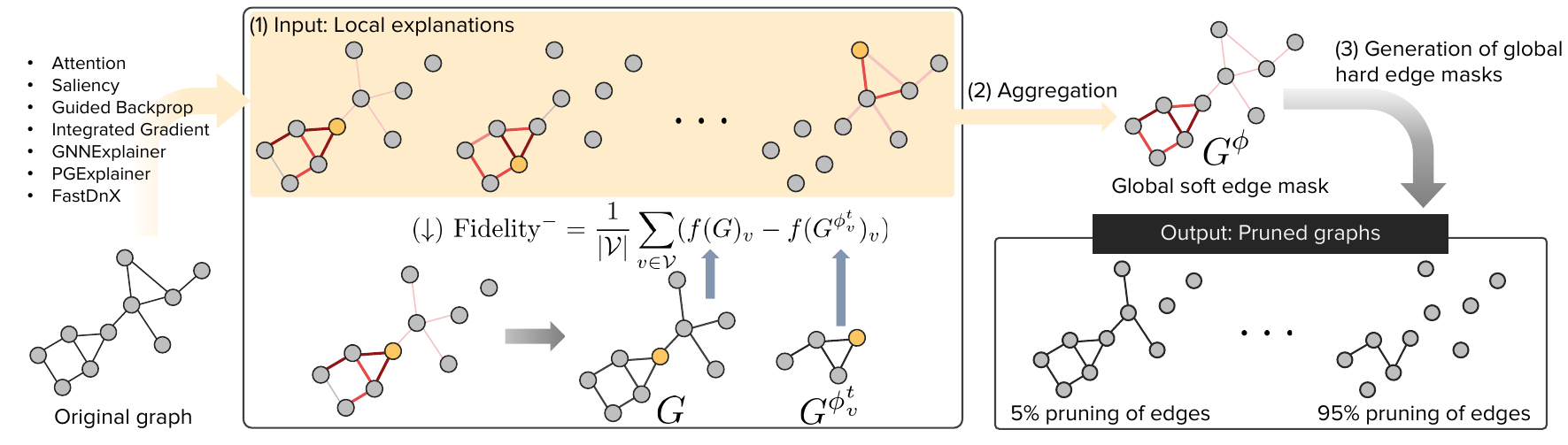}
    \caption{The overview of the FiP framework.}
    \label{fig:fidelitypruningoverview}
    \vskip -0.2in
\end{figure*}

In this work, we investigate the feasibility of this hypothesis, i.e., we are interested in using the intuition of fidelity itself in the context of pruning the edges from the input graph. In other words, we attempt to use the aggregate of given local explanations for each node for pruning edges from the input graph. Towards this end, we first provide a simple yet effective framework, built upon any edge attribution methods, to prune the input graph based on local explanations. Our empirical results from 7 different edge attribution methods comprehensively demonstrate the feasibility of using XAI methods for graph pruning. Surprisingly, we find that explanation methods that are specifically designed for GNN methods does not perform well in graph pruning, although they have superb performance in fidelity$^{-}$. Our analysis further validates this finding by explicitly visualizing pruned graphs, while emphasizing the necessity of developing a more sophisticated aggregation method.

\section{Related Work}
\paragraph{Fidelity.} The fidelity metric, i.e., the measurement on the subset of input features highlighted by the explanation for the actual relevance to the model, has been acknowledged as one of the core properties for explanation~\cite{Yeh2019fidelity}. In GNN explanations, two variants, i.e., fidelity$^{+}$ and fidelity$^{-}$, are widely used depending on the unimportant/important part to be removed from the input~\cite{Yuan2023GNNXAISurvey}. Although the fidelity measures are empirical measurements, theoretical analysis on its robustness~\cite{Agarwal2022fidelity,Zheng2024fidelity} has also been performed.
\paragraph{Graph pruning.} Removing irrelevant edges from the underlying graph is a common strategy for reducing the computation complexity of the GNN models, as the complexity is known to be proportional to the number of edges~\cite{Chiang2019clustergcn}. Different ways to identify such irrelevant edges, including training a separate neural network~\cite{Zheng2020neuralsparse}, using effective resistance measures~\cite{Srinivasa2020EffectiveResistance,liu2023dspar} and using graph lottery ticket hypothesis~\cite{Liu2023LTHPruning}, were presented.

Our work focuses on the unique task of utilizing the intuition of the fidelity measurement on graph explanations to be used to prune edges from the input graph. To the best of our knowledge,~\cite{Naik2024iterativeenhancement} lies in a similar objective to our study; however, it focuses on providing additional node features as a result.

\section{Methodology}
\subsection{Basic Notations and Problem Settings}
We denote an undirected and unweighted graph as $G=(\mathcal{V}, \mathcal{E}, X, \mathcal{A})$, where $\mathcal{V}$ is the set of nodes, $\mathcal{E} \subseteq \mathcal{V} \times \mathcal{V}$ is the set of edges, $X \in \mathbb{R}^{|\mathcal{V}| \times d}$ is the node feature matrix, and $\mathcal{A}: \mathcal{E} \rightarrow \mathbb{R}$ maps each edge in $\mathcal{E}$ to a real number (representing the set of edge weights or attributions). Also, we denote the set of neighbors for node $v$ as $\mathcal{N}_v$. We focus on node classification, where a set of classes $C = \{1, ... ,c\}$ are given. Then, we denote the one-hot label matrix $Y \in \{0, 1\}^{|\mathcal{V}| \times |\mathcal{C}|}$, where $Y_{v, :} = {\bf y}_v$ is the ground-truth label for node $v$. We assume that we are given a pre-trained $L$-layered GNN model $f$, which produces a prediction $\hat{Y} \in \mathbb{R}^{|\mathcal{V}| \times |\mathcal{C}|}$.

We consider a GNN explanation method $\Phi$ that takes a target node $v$, a target output $t \in \mathcal{C}$ as input and assigns a non-negative edge attribution value to a given edge $e_{i,j}$ on the GNN model, i.e., $\Phi(e_{i,j}; v, t) \coloneqq \phi_{v}^t(i,j) \in \mathbb{R}$. We set $t$ as the predicted class of $v$ unless otherwise stated.\footnote{Although explanation methods often provide feature masks, we focus on edge-wise explanations in this work.} By collecting the edge attribution values $\phi_{v}^t(i,j)$ for node $v$, we can construct a soft mask over the input graph, denoted as $G^{\phi_v^t}$.  

\begin{figure*}
    \centering
    \includegraphics[width=0.9\textwidth]{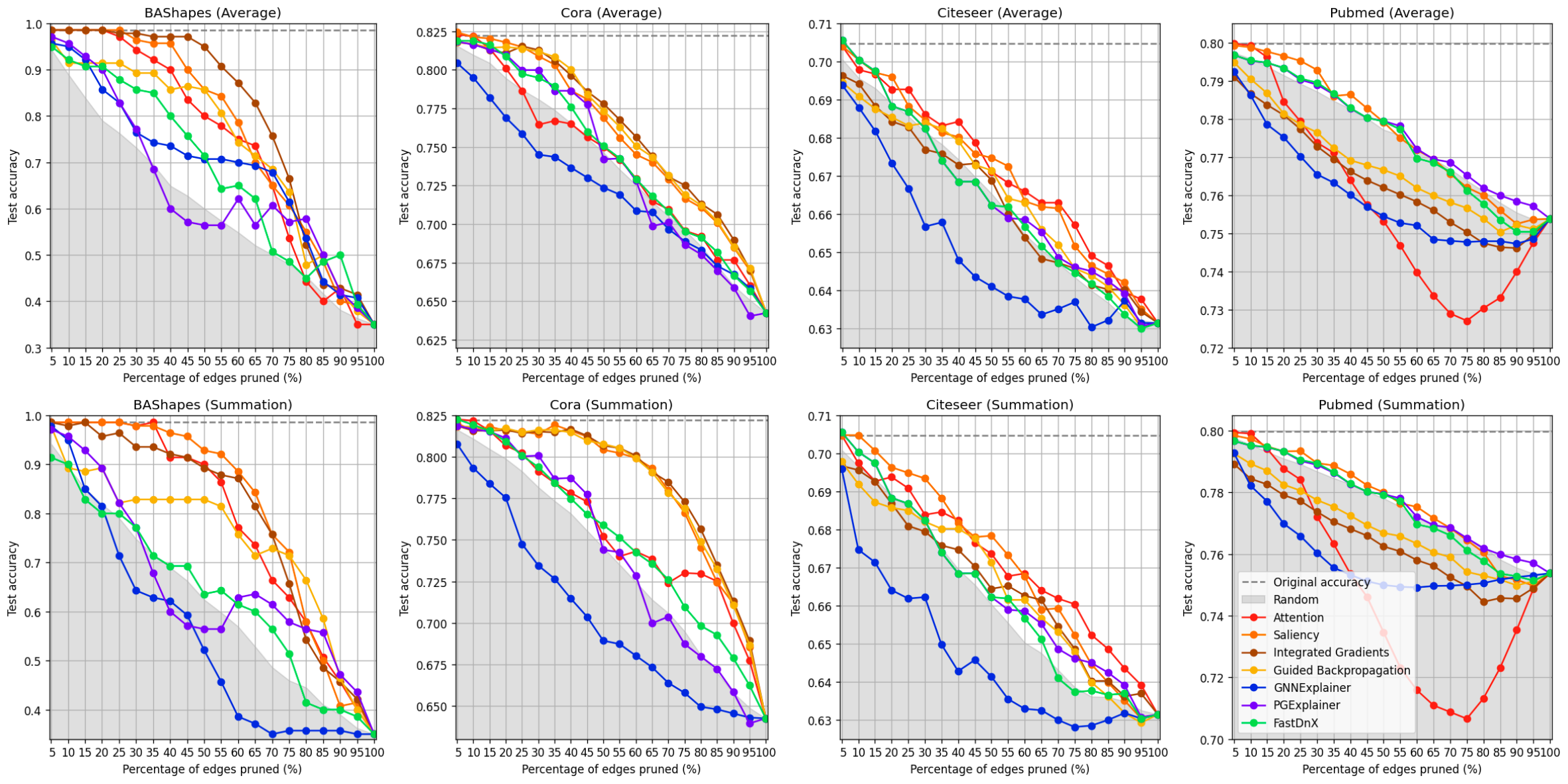}
    \caption{Graph pruning performance of FiP for 7 edge attribution methods as well as a random baseline on 4 benchmark datasets. The grey area indicates the performance of random attributions, and the dashed line indicate the test performance without any pruning.}
    \label{fig:maingraphpruningexp}
    \vskip -0.1in
\end{figure*}

\subsection{Fidelity$^{-}$-inspired Pruning Framework}
To utilize the local explanations for graph pruning, we propose \textbf{F}idelity$^-$-\textbf{i}nspired \textbf{P}runing, \textbf{FiP}, a simple yet effective framework that aggregates the edge attribution scores and creates a global edge mask (see Figure~\ref{fig:fidelitypruningoverview}). The step-by-step description of FiP is as follows:
\begin{enumerate}
    \item Explanations (i.e., local soft edge masks $G^{\phi_v^t}$) for each target node (yellow nodes in the figure) obtained by a specific edge attribution method are taken as input.
    \item The local soft edge masks $G^{\phi_v^t}$ are aggregated over all $v \in \mathcal{V}$ via summing or averaging the edge attributions to generate $G^{\phi}$ (i.e., turning $\phi_{v}^t(i,j)$ into a global soft edge mask $\phi(i,j)$).
    \item Hard edge masks are generated via discarding edges with the lowest aggregated edge attribution scores $\phi(i,j)$.
\end{enumerate}

We can expect that the performance gradually decreases when we prune more edges (eventually resulting in information loss of the input), but a \textit{good} global soft edge mask $G^{\phi}$ will assign a low score to noisy edges and a higher score to edges that severely hurt the performance when removed. Note that the process of FiP can be interpreted as a global version of fidelity$^-$, since both FiP and fidelity$^-$ discard unimportant edges in $G^{\phi_v^t}$ or $G^{\phi}$. Although there may be more sophisticated methods to aggregate local edge attributions aside summation and averaging them, we leave the design of such methods as future work.

\section{Empirical Observations}
In this section, we evaluate the graph pruning performance of FiP using various GNN explanation methods.

\subsection{Basic Settings}

In our experiments, we train a 2-layer GAT model~\cite{Velickovic2018GAT} on 4 benchmark datasets, BA-Shapes~\cite{Ying2019GNNExplainer}, Cora, Citeseer, and Pubmed~\cite{Yang2016Planetoid}, where the model achieves test performance of 0.9857, 0.8531, 0.7389, and 0.8056, respectively. As mentioned, we only consider average or summation when aggregating local edge attributions in FiP.

\subsection{Explanation Methods}
We adopt the following seven edge attribution methods commonly used in the literature.
\paragraph{Attention (Att).} The edge attention weights are treated as a proxy of edge attribution. We average the attention weights over all layers, similarly as in~\cite{Ying2019GNNExplainer,SanchezLengeling2020gnnxaieval}.
\paragraph{Saliency (SA).}~\cite{Simonyan2013Saliency} is the absolute value of the gradient with respect to the input.
\paragraph{Integrated Gradient (IG).}~\cite{Sundararajan2017IntegratedGradient} calculates an edge attribution score via approximating the integral of gradients of the model’s output with respect to the input along the path from a baseline to the input.
\paragraph{Guided Backpropagation (GB).}~\cite{Springenberg2014GuidedBackprop} is similar to SA, except that the negative gradients are clipped during backpropagation, basically focusing on features with an excitation effect.
\paragraph{GNNExplainer (GNNEx).}~\cite{Ying2019GNNExplainer} is the most widely used explanation method tailored to GNNs, where it identifies a local subgraph most relevant to the model's predictions by maximizing the mutual information.
\paragraph{PGExplainer (PGEx).}~\cite{Luo2020PGExplainer} trains a separate parameterized mask predictor to generate edge masks that identify edges important to the prediction. 
\paragraph{FastDnX (FDnX).}~\cite{Pereira2023FastDnX} is a recently proposed method for explaining GNNs, where it basically relies on a surrogate SGC model~\cite{Wu2019SGC} to explain the model's behavior.

\begin{figure*}[t]
    \centering
    \includegraphics[width=\textwidth]{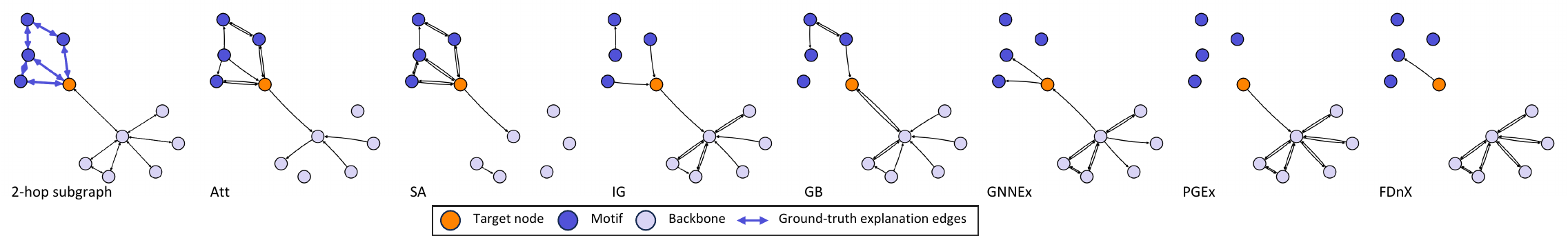}
    \caption{Visualizations of graph pruning using different edge attribution methods by removing 50\% of the edges from the original graph.}
    \label{fig:CaseStudy}
    \vskip -0.2in
\end{figure*}

\subsection{Experimental Results on Graph Pruning}
We show the experimental results of using various explanation methods in FiP with our findings. As our main result, Figure~\ref{fig:maingraphpruningexp} shows the test performance when we remove $p$\% of the edges with the lowest global soft edge mask $G^\phi$, where $p \in \{5, 10, \cdots, 100\}$. For comparison, we also show the performance when we use random attributions (averaged over 10 independent trials), depicted as the grey area. From Figure~\ref{fig:maingraphpruningexp}, we make the following observations:
\begin{itemize}
    \item Overall, edge attribution methods reveal their potential in edge pruning. For example, on BAShapes, we can delete half of the edges using IG only with the performance degradation of less than $4\%$.
    \item Despite being rare, there are cases where the accuracy after pruning outperforms the original test accuracy (e.g., on Citeseer, 5\% pruning with FastDnX by summation).
\end{itemize}
Table~\ref{tab:AverageRank} summarizes the rank in performance averaged over 20 different pruning cases for 7 edge attribution methods (as well as a random baseline) on 4 benchmark datasets. Our findings are as follows.
\begin{itemize}
    \item The best edge attribution method for graph pruning tends be one of Att, SA, and IG. This is quite unexpected, as SA and IG are `general' XAI methods (i.e., the ones not tailored to GNNs). In a similar regard, GNNExplainer tends to exhibit the worst performance for most of the datasets.
    \item There are no significant differences between using summation and average for aggregating local edge attributions in FiP.
\end{itemize}

\begin{table}[t]
\small
\centering
\resizebox{\columnwidth}{!}{
    \begin{tabular}{lcccc}
    \toprule
    Method & BAShapes & Cora & Citeseer & Pubmed \\
    \midrule
    Att & 3.63/2.26 & 4.89/4.11 & \textbf{1.74}/\textbf{1.89} & 6.32/6.58 \\
    SA & 2.42/\textbf{1.58} & 2.58/2.47 & 1.89/2.11 & \textbf{2.21}/\textbf{2.11} \\
    IG & \textbf{1.53}/2.58 & \textbf{1.95}/\textbf{1.84} & 5.32/4.26 & 6.58/6.58 \\
    GB & 3.84/3.68 & 2.16/2.26 & 4.42/5.05 & 5.16/5.42 \\
    GNNEx & 5.11/7.42 & 7.42/7.95 & 7.68/7.79 & 7.26/6.74 \\
    PGEx & 5.58/4.84 & 5.84/5.58 & 4.00/3.84 & \textbf{2.21}/2.42 \\
    FDnX & 5.32/6.16 & 4.53/4.42 & 4.68/4.53 & 3.16/3.05 \\
    Random & 7.53/6.47 & 6.42/6.68 & 5.47/5.68 & 3.05/3.00 \\
    \bottomrule
   \end{tabular}
   }
  \caption{Rank in performance averaged over 20 different pruning percentages (average/summation).}
  \vskip -0.2in
  \label{tab:AverageRank}
\end{table}

\subsection{Visualizations of Graph Pruning}
We observe the effect of using different edge attributions for graph pruning by visualizing the resulting graphs. Figure~\ref{fig:CaseStudy} visualizes results on BA-Shapes for a specific target node (yellow nodes) using FiP with summation. We choose BA-Shapes since it is the only dataset that contains ground-truth explanation edges (see blue arrows). Note that only ground-truth explanation edges are meaningful in the sense that they construct house-shapes (i.e., motifs), influencing the ground-truth node labels included in the motif (dark blue and yellow nodes). The remaining edges do not have any semantic meanings and merely serve as a backbone structure of the graph. By setting $p=50$ (i.e., removing half of the edges), we observe the following:
\begin{itemize}
    \item Edge attribution methods showing superior performance on BA-Shapes (i.e., Att, SA, IG, and GB) tend to prune edges that are not included in the ground-truth explanation edges compared to GNNex, PGEx, and FDnX.
    \item Especially for Att and SA, the resulting graphs after pruning tend to be less noisy and explainable.
\end{itemize}

\begin{table}[t]
\small
\centering
\resizebox{\columnwidth}{!}{
    \begin{tabular}{lcccc}
    \toprule
    Method & BAShapes & Cora & Citeseer & Pubmed \\
    \midrule
    Att & $4.06 \times 10^{-2}$ & $3.67 \times 10^{-2}$ & $2.23 \times 10^{-2}$ & $2.46 \times 10^{0}$ \\
    SA & $3.54 \times 10^{-7}$ & $2.21 \times 10^{-7}$ & $\mathbf{8.90 \times 10^{-8}}$ & $2.46 \times 10^{0}$ \\
    IG & $6.25 \times 10^{0}$ & $1.26 \times 10^{0}$ & $5.68 \times 10^{-1}$ & $\mathbf{2.25 \times 10^{0}}$ \\
    GB & $3.77 \times 10^{0}$ & $1.42 \times 10^{0}$ & $7.04 \times 10^{-1}$ & $2.40 \times 10^{0}$ \\
    GNNEx & $\mathbf{3.44 \times 10^{-7}}$ & $\mathbf{2.14 \times 10^{-7}}$ & $3.52 \times 10^{-1}$ & $2.46 \times 10^{0}$ \\
    PGEx & $3.83 \times 10^{-7}$ & $2.04 \times 10^{-2}$ & $7.11 \times 10^{-3}$ & $2.46 \times 10^{0}$ \\
    FDnX & $1.41 \times 10^{-1}$ & $1.77 \times 10^{-2}$ & $7.05 \times 10^{-3}$ & $2.46 \times 10^{0}$ \\
    \bottomrule
   \end{tabular}
   }
  \vskip -0.1in
  \caption{Measurement of fidelity$^{-}$.}
  \label{tab:FidelityExperiment}
\end{table}

\subsection{Relationship with Fidelity Scores}
We perform further analysis by measuring the average fidelity$^{-}$ scores over all nodes in the graph for each edge attribution method. In our setting, we measure fidelity$^{-}$ as the average output (logit) difference between using the original graph and a sparser graph as input, where the sparse graph is generated by removing 50\% of the edges with the lowest edge attribution scores in $G^{\phi}$. Lower fidelity$^{-}$ indicates a higher quality explanation for each node. Table~\ref{tab:FidelityExperiment} summarizes the measurement of fidelity$^{-}$ scores for all edge attribution methods and datasets. Here, we find that the fidelity$^{-}$ does not necessarily translate to graph pruning performance. As an example, GNNEx shows the best performance in fidelity$^{-}$ for the Cora dataset; however, the average rank when using GNNEx in FiP is 5.84 and 5.58 for average and summation aggregation cases, respectively.

\section{Discussion and Conclusion}
In this work, we have empirically validated the feasibility of using local edge attribution methods for edge pruning. Our empirical analysis demonstrated that 1) local edge attributions can be effectively used for graph pruning with our FiP framework and 2) general XAI methods outperform XAI methods tailored to GNN models. We believe that FiP will not only improve efficiency but also eventually result in sparser explanations, which makes manual inspection feasible~\cite{Pope2019sparseexplanation}, as sparse explanations are generally considered to be more human-comprehensible~\cite{Yuan2023GNNXAISurvey,Funke2023Zorro}. Potential avenues of future work include development of more sophisticated aggregation methods as well as investigation of the relationship between sparsity levels and human comprehensiveness on the explanations.

\section*{Acknowledgements}
This research was supported by the National Research Foundation of Korea (NRF) grant funded by the Korea government (MSIT) (No. 2021R1A2C3004345, No. RS-2023- 00220762).

\bibliographystyle{named}
\bibliography{ijcai24}

\end{document}